\pdfoutput=1
\documentclass[11pt]{article}

\usepackage[preprint]{acl}
\usepackage{times}
\usepackage{amsmath}
\usepackage{latexsym}
\usepackage{inconsolata}
\usepackage[T1]{fontenc}
\usepackage{xcolor}
\usepackage{hyperref}
\usepackage{microtype}
\usepackage{arydshln}
\usepackage{inconsolata}

\usepackage{graphicx}
\newcommand{\myparagraph}[1]{\vspace{0.2em}\noindent\textbf{#1}}
\usepackage{multirow}
\usepackage{booktabs,paralist}
\usepackage{algorithm}
\usepackage[noend]{algpseudocode}

\title{Not Every Token Needs Forgetting: Selective Unlearning to Limit Change in Utility in Large Language Model Unlearning}

\author{\textbf{Yixin Wan\textsuperscript{1,2}},
  \textbf{Anil Ramakrishna\textsuperscript{2}},
  \textbf{Kai-Wei Chang\textsuperscript{1,2}},
  \textbf{Volkan Cevher\textsuperscript{2}},
  \textbf{Rahul Gupta\textsuperscript{2}}
  \\
{
\textsuperscript{1}University of California, Los Angeles,
 \textsuperscript{2}Amazon
 }
\\
 {
   \texttt{elaine1wan@g.ucla.edu}
 }
}

\begin{document}

\maketitle

\begin{abstract}
% \vspace{-0.5em}
% Large Language Model (LLM) unlearning has recently gained significant attention, driven by the need to remove unwanted information—such as private, sensitive, or copyrighted content—from trained models. In conventional approaches, given a document to be forgotten, LLMs perform unlearning by updating parameters to minimize negative language model loss on all tokens in that document. However, frequent words and general concepts (e.g., pronouns, prepositions, common nouns) should not be unlearned. In this paper, we propose Selective Unlearning (SU), which identifies a critical subset of tokens within the forgetting set that is relevant to the unwanted information and unlearns only those tokens. Extensive experiments on two benchmarks and six baseline unlearning algorithms demonstrate that SU not only achieves effective unlearning on the targeted forget data, but also significantly preserves the model’s utility in the retaining set.
Large Language Model (LLM) unlearning has recently gained significant attention, driven by the need to remove unwanted information, such as private, sensitive, or copyrighted content, from LLMs. However, conventional unlearning approaches indiscriminately update model parameters to forget all tokens in a target document, including common tokens (e.g., pronouns, prepositions, general nouns) that carry general knowledge. In this paper, we highlight that ``not every token needs forgetting''.
We propose \textbf{Selective Unlearning (SU)}, which identifies a critical subset of tokens within the forgetting set that is relevant to the unwanted information, and unlearns only those tokens. 
% , a strategy that leverages two assistant models with different knowledge scopes to identify tokens uniquely associated with the content to forget. SU then unlearns only this critical subset of tokens, mitigating interference with general knowledge and preserving the model’s overall utility. 
Experiments on two benchmarks and six baseline unlearning algorithms demonstrate that SU not only achieves effective unlearning on the targeted forget data, but also significantly preserves the model’s utility in the retaining set.
% \vspace{-0.5em}
\end{abstract}

\section{Introduction}
Text corpora used to train Large Language Models (LLMs) often contain sensitive, private, or copyrighted content. To address the risks posed by such data, recent research has explored \textit{LLM unlearning}---aims to remove specific unwanted knowledge from a model without incurring the cost and effort of retraining from scratch.
% Text documents used to train LLMs often include sensitive, private, or copyrighted information. To address the risks associated with these data, a growing body of research focuses on ``unlearning'' in LLMs, which aims to remove specific unwanted knowledge from a model without incurring the cost and effort of retraining from scratch.

\begin{figure}[t]
\vspace{-1em}
% \hspace*{-0.5cm}  
\includegraphics[width=0.47\textwidth]{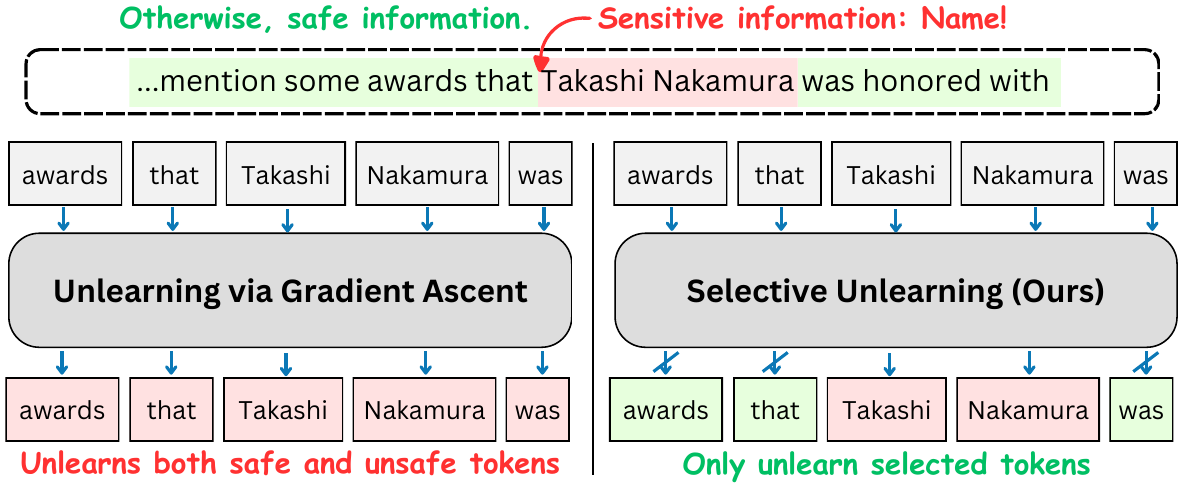}
\vspace{-0.5em}
    \caption{\small \label{fig:example_selection} Example of how tokens are selected for unlearning. Red blocks indicate unlearned tokens, on which the forgetting loss is calculated. SU avoids the forgetting of general information like ``that'', therefore preserving model utility.}
%\vspace{-1.5em}
\end{figure}

Existing unlearning approaches typically apply the same unlearning loss to every token in the targeted documents. 
However, as illustrated in Figure~\ref{fig:example_selection},
this approach forces the model to unlearn not only sensitive information but also general concepts.
%unintentionally reducing the model’s overall utility. 
%As illustrated in Figure~\ref{fig:example_selection}, 
Even benign tokens like ``that'' or ``she'' in the target forget documents are unlearned, unnecessarily degrading the model’s language capabilities.
% Existing unlearning methods often apply the same unlearning loss to every token in the targeted documents. However, this approach forces the model to unlearn general concepts, it can unintentionally reduce the model’s overall utility. 
% For instance, as Figure \ref{fig:example_selection} demonstrates,  the target document contains general tokens such as ``that'' or ``she.'' While these tokens do not carry specific private information, they encode essential linguistic knowledge. Therefore, unlearning them is unnecessary and may degrade the model’s language understanding.

Motivated by this, we contend that \emph{\textbf{not every token needs forgetting}}: %models should \emph{not} be indiscriminately stripped of knowledge related to general tokens, and 
an unlearning method should selectively target only tokens that encode unique information in the forget set.
% We propose the \textbf{Selective Unlearning (SU)} strategy, a novel framework that utilizes 2 assistant models with different scopes of knowledge to identify tokens containing information unique to the forget set within instances, and unlearn only these specific subsets of tokens.
To this end, we introduce \textbf{Selective Unlearning (SU)}, a novel framework that utilizes two assistant models with different scopes of knowledge to identify and unlearn only a subset of tokens that carry forget-specific information. 
By only calculating unlearning losses on these tokens with forget set-specific information, SU can reduce unnecessary interference with retained information, thereby preserving model utility on general knowledge.

%To evaluate the efficacy of SU, 
We conduct extensive experiments on 2 popular benchmarks: Task of Fictitious Unlearning (TOFU)~\citep{maini2024tofu} and MUSE-News~\citep{shi2024muse} to compare SU with 6 unlearning methods. 
Results demonstrate that SU not only achieves comparable unlearning quality to the existing methods, but also substantially improves the preservation of retained knowledge. 
Striking a balance between unlearning and utility preservation, SU represents a promising step toward scalable and utility-preserving unlearning strategies for LLMs.
% These findings highlight SU as a promising direction for scalable and utility-preserving unlearning in LLMs.

%\vspace{-0.1cm}
\section{Related Work}
% \kw{We need to discuss the relation with "Rho-1: Not All Tokens Are What You Need".}
\subsection{Unlearning for LLMs}
Previous works on unlearning have explored ways to remove sensitive, private, or copyrighted information~\citep{carlini2021extracting} from LLMs. 
The most intuitive method is \textbf{Gradient Ascent (GA)}~\citep{jang2023knowledge, yao2023large}, which maximizes the language model loss\footnote{Equivalently, it minimizes the negative language model loss.} on the forget dataset.
However, GA has been shown to degrade the performance of models in data and knowledge outside of the forget set, even resulting in model collapsing~\citep{zhang2024negative}.

With this in mind, prior studies have proposed ways to better preserve model performance on retain data.
% regularize GA unlearning as well as alternative methods to 
For instance, researchers have proposed to apply gradient descent~\citep{liu2022continual, maini2024tofu} or regularize models' KL-divergence~\citep{wang2024rkld, chen2023unlearn} on the retain set during unlearning.
The former is also known as \textbf{``Gradient Difference (GD)''}, since it essentially optimizes the difference between losses on forget and retain data.
Additionally, previous research also investigated alternatives to the GA approach, with \textbf{Negative Preference Optimization (NPO)}~\citep{zhang2024negative} being one of the most promising algorithms.
NPO uses forget candidates as negative examples in Direct Preference Optimization (DPO)~\citep{rafailov2024direct}, avoiding model collapse.
To better assess different unlearning algorithms, more recent works construct LLM unlearning benchmarks such as TOFU~\citep{maini2024tofu},  MUSE~\citep{shi2024muse} and LUME~\citep{ramakrishna2025lume,ramakrishna2025semevaltask4}.

\subsection{Selecting Unlearning Candidates}
Although previous research on unlearning in LLMs has achieved remarkable progress, most of them formulate the task as such that models must be retrained to remove information about all candidates in the forget set. 
Most related to the
work, \citet{wang2024selectiveforgettingadvancingmachine} proposed to unlearn parts in a sequence that has lower log-probability than a threshold.
% Results show that the selection mechanism is able to: (1) ensure model robustness when real-world knowledge is injected into unlearning data, and (2) better retain the real-world common sense information in models.
However, their experiments were limited to variations of the GPT-Neo model~\citep{gao2020pile}, and were not extended to the newer LLMs.
\citet{ma2024unveilingentitylevelunlearninglarge} and ~\citet{choi2024optoutinvestigatingentitylevelunlearning} explored entity-level unlearning, which selectively unlearns knowledge related to specific entities, instead of all knowledge in the forget set.
~\citet{mccartney2024introducing} selectively chooses anti-knowledge, or knowledge that conflicts with a model's original memory, for unlearning.
Similarly, ~\citet{choi2024optoutinvestigatingentitylevelunlearning} proposed to utilize a LLM trained with negative instructions to produce obliterated generations for unlearning.
However, these approaches still require forgetting full chunks of text, among which common words and tokens inevitably persist.
% Consequently, they still lack the ability to preserve utility by retaining models' understanding of universal tokens.

In the field of language model pre-training, ~\citet{lin2024rho}'s work showed that \textbf{not all tokens are needed} for training a model.
Specifically, they used a reference model for scoring different tokens in training data, and calculated a focused loss specifically on tokens with higher scores.
Inspired by their method, we design a token-level selection strategy that utilizes 2 assistant models with different knowledge, which specifically targets the unlearning task.

\section{Selective Unlearning} % methodology
\begin{figure}[t]
\vspace{-0.5em}
% \hspace*{-0.5cm}  
\includegraphics[width=0.49\textwidth]{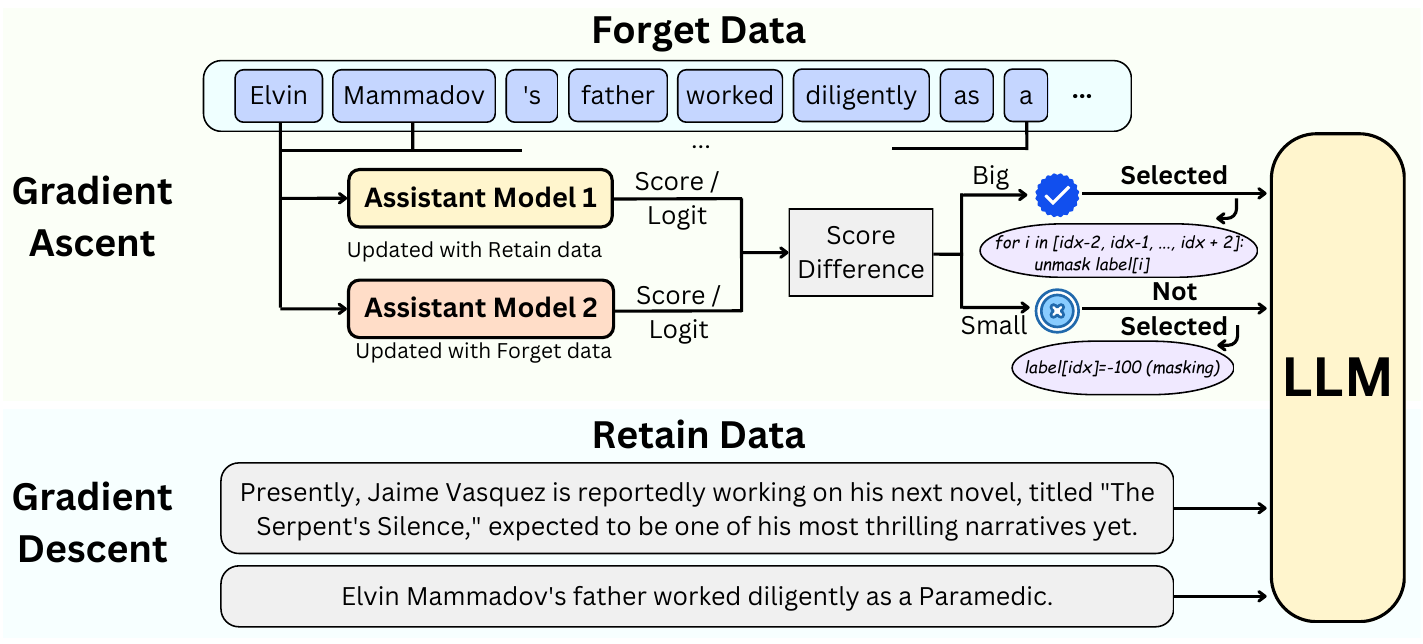}
\vspace{-1.5em}
    \caption{\small \label{fig:tlsu} The proposed SU framework. We use 2 assistant models, trained on different data splits, to facilitate the token selection process. Based on the difference between their prediction scores, we can choose to only unlearn tokens that contain information unique to the forget dataset.}
\vspace{-1em}
\end{figure}

% We propose \textbf{Selective Unlearning}, which involves selectively updating the model using only a subset of all unlearning candidates.
% Specifically, w
We introduce \textbf{Selective Unlearning (SU)}, which selectively unlearns a subset of tokens with information unique to the forget set. 
We apply SU to do selective Gradient Ascent on models, while at the same time using Gradient Descent on all retain data to better preserve model performance.
% This section discusses our methodology to construct the token selection mechanism.
Figure \ref{fig:tlsu} provides an overview of our SU framework.

% \subsection{Methodology}
% \label{subsec:methodology}
\subsection{Selection Criteria Construction}
% Assistant Model Training
SU adopts a selection mechanism to only unlearn tokens that contain unique information for the forgotten set.
To identify which tokens possess forget data-unique information, we introduce two assistant models to construct the selection criteria.
The two models are trained with different data splits, and 
% Only one of the 2 models was trained on the forget data.
therefore only possess knowledge of different proportions of data (e.g., one model has knowledge of full data, another only knows retain data).
We can then use the behavior divergence between the models to identify forget data-specific tokens.
Specifically, SU selects unlearn tokens by placing a threshold on the difference between the prediction scores or logits of the two models.
% The two assistant models are trained on different splits of data, and the difference between their prediction scores is used to identify tokens with forget set-specific information.
Table \ref{tab:assistant-data} summarizes selection criteria for assistants trained with different combinations of splits.

For instance, for a model \(f_{\theta}\) that memorizes a sequence \(t\) with n tokens \(t_1, t_2, ..., t_n\), let one assistant model \(f_{\theta}^1\) be trained on full data and another \(f_{\theta}^2\) on retain data.
Let \(\gamma\) be the selection threshold.
For a token \(t_i\), let \(S(\cdot )\) denote a selection function with ``1'' meaning selected and ``0'' meaning not selected for unlearning.
% the selection function indicating whether \(t_i\) would be chosen for unlearning can be denoted as:
Then,
\vspace{-0.3em}
\begin{equation*}
\small
\vspace{-0.2em}
    S(t_i) = \left\{\begin{array}{l}
    1,\; \text{if } |p_{\theta}^1(t_i | t_{<i}) - p_{\theta}^2(t_i | t_{<i})| > \gamma; \\
    0,\; \text{otherwise.} \\
    \end{array}
    \right.
\vspace{-0.2em}
\end{equation*}

% SU adopts a selection mechanism to only unlearn tokens that contain unique information for the forgotten set.
% To identify which tokens possess forget data-unique information, we introduce two assistant models to construct the selection criteria.
% The models are trained with different data splits, and 
% % Only one of the 2 models was trained on the forget data.
% therefore only possess knowledge of different proportions of data (e.g., one model has knowledge of full data, another only knows retain data).
% We can then use the behavior divergence between the models to identify forget data-specific tokens.
% Specifically, SU selects unlearn tokens by placing a threshold on the difference between the prediction scores or logits of the two models.

% Specifically, for a model \(f_{\theta}\) that memorizes a sequence \(t\) with n tokens \(t_1, t_2, ..., t_n\), t
The original GA algorithm unlearns \(t\) by maximizing the language model loss:
\vspace{-0.5em}
\begin{equation*}
\small
\vspace{-0.5em}
    \mathcal{L}_{GA}(f_{\theta}, t) = -\sum\nolimits_{i=1}^n \log (p_{\theta} (t_i | t_1, ..., t_{i-1}))
\vspace{-0.3em}
\end{equation*}
% \vspace{-0.3em}
% For SU, only the unlearning loss for 5-grams surrounding each selected token is calculated as shown in Algorithm \ref{alg:1}.
, in which \(p_{\theta}\) represents the output probability. As shown in Algorithm \ref{alg:1}, we calculate the unlearning loss for 5-grams surrounding each selected token to ensure the removal of complete information related to the token. Our preliminary experiments show that this helps to remove the whole phrase surrounding the token.
% Let \(S(\cdot )\) denote a selection function, where:
% \vspace{-0.2em}
% \begin{equation}
% \small
% \vspace{-0.2em}
%     S(t_i) = \left\{\begin{array}{l}
%     1,\; \text{if token \(t_i\) is selected}\\
%     0,\; \text{otherwise.} \\
%     \end{array}
%     \right.
% \vspace{-0.3em}
% \end{equation}
% \vspace{-0.3em}
% Then, the SU loss can be calculated as:

\begin{algorithm}
\vspace{-0.2em}
\small
\caption{\label{tlsu-algo}Calculating SU loss.}
\label{alg:1}
\begin{algorithmic}[1]
    \State \textbf{Part 1}
    \State Initialize an empty list for storing selected token positions \(l=\) [].
    \For{$i \in [1, 2, ..., n]$}
        \State $sel_i = S(t_i)$ \Comment{Whether token \(t_i\) is selected for unlearning}
        \If{$sel_i == 1$} \Comment{Selected}
            % \State $i \gets i-1$
            \For{$j \in [i-2, i-1, i, i+1, i+2]$}
                \State Add $j$ to $l$
            \EndFor
        \ElsIf{$i \in l$} \Comment{Not Selected, no loss calculated}
            \State Remove $i$ from $l$
        \EndIf 
    \EndFor \\
    \hrulefill
    \State \textbf{Part 2}
    \State Initialize unlearning loss $\mathcal{L}_{SU}=0$.
    \For{$idx \in l$} \Comment{Indexes of tokens to calculate loss on}
        \State $\mathcal{L}_{SU} += (-\log (p_{\theta} (t_{idx} | t_1, ..., t_{idx-1})) $
    \EndFor \\
    \Return $\mathcal{L}_{SU}$
\end{algorithmic}
\end{algorithm}
\vspace{-1.5em}

\subsection{Implementation}
%There are different implementation-level variations of our SU framework.
% We discuss them in the following section.
%\subsubsection{Assistant Model Structure}
% \myparagraph{N-Gram Model as Selection Assistant}
% \myparagraph{LLM as Selection Assistant}
We experiment with two different model structures for the selection assistant models.
% : Statistical Language Models and Neural Language Models.

\myparagraph{Statistical: N-Gram Language Models}~\citep{brown-etal-1992-class} learn and predict the probability of ``N-grams''---or continuous sequences of n tokens---in texts.
We experiment with N-Gram models due to their efficiency and interoperability.
In Appendix \ref{sec:appendix-ablation} Table \ref{tab:ngram-memory}, we demonstrate the memory efficiency of N-Gram-based assistant models---even trained on full data, the model only takes around 20M of memory.

\myparagraph{Neural: LLMs} adopt neural-based structures that learn to capture meanings and relationships between language features in latent space.
We experiment with LLMs due to their outstanding language understanding abilities.

% \subsubsection{Selection Criteria Construction}
% % Assistant Model Training
% The two assistant models are trained on different splits of data, and the difference between their prediction scores is used to identify tokens with forget set-specific information.
% Table \ref{tab:assistant-data} summarizes selection criteria for assistants trained with different combinations of splits.

% For instance, let one assistant model \(f_{\theta}^1\) be trained on full data and another \(f_{\theta}^2\) on retain data.
% Let \(\gamma\) be the selection threshold.
% For a token \(t_i\), the selection function indicating whether \(t_i\) would be chosen for unlearning can be denoted as:
% \vspace{-0.3em}
% \begin{equation}
% \small
% \vspace{-0.2em}
%     S(t_i) = \left\{\begin{array}{l}
%     1,\; \text{if } |p_{\theta}^1(t_i | t_{<i}) - p_{\theta}^2(t_i | t_{<i})| > \gamma; \\
%     0,\; \text{otherwise.} \\
%     \end{array}
%     \right.
% \vspace{-0.2em}
% \end{equation}
% % when one assistant model is trained on the full dataset and another on the retain data, we only choose to unlearn tokens for which the difference between the prediction scores of the 2 assistants are \textbf{greater} than a threshold.

\begin{table}[t]
\centering
\scriptsize
\vspace{-0.2em}
\begin{tabular}{p{0.07\textwidth}p{0.07\textwidth}p{0.245\textwidth}}
\toprule
\multicolumn{2}{c}{\textbf{Training Data Split}}  &  \multirow{2}*{\textbf{Selection Criteria}} \\
\cmidrule{1-2}
\textbf{Assistant 1}   & \textbf{Assistant 2}  &  \\
\midrule
Full & Retain & Score difference \textbf{greater than} threshold. \\
Full & Forget & Score difference \textbf{smaller than} threshold. \\
Retain & Forget & Score difference \textbf{greater than} threshold. \\
\bottomrule
\end{tabular}
\vspace{-0.5em}
\caption{Combinations of data splits for training assistant models, and corresponding selection criteria.}
\label{tab:assistant-data}
\vspace{-1.5em}
\end{table}

\vspace{-0.6em}
\section{Experiments}
\vspace{-0.3em}
We demonstrate the effectiveness of SU through experiments on 6 baselines and 2 benchmarks.

\begin{table*}[t]
\vspace{-1em}
\centering
\small
\begin{tabular}{p{0.145\textwidth}p{0.07\textwidth}p{0.12\textwidth}p{0.1\textwidth}p{0.07\textwidth}p{0.07\textwidth}p{0.1\textwidth}p{0.1\textwidth}}
\toprule
\multirow{3}*{\textbf{Method}} & \multicolumn{3}{c}{\textbf{MUSE}}  & \multicolumn{4}{c}{\textbf{TOFU}} \\
\cmidrule(lr){2-4} \cmidrule(lr){5-8}
& \multicolumn{2}{c}{\textbf{Forget}} & \multicolumn{1}{c}{\textbf{Utility}} &  \multicolumn{1}{c}{\textbf{Forget}} & \multicolumn{3}{c}{\textbf{Utility}} \\
\cmidrule(lr){2-3} \cmidrule(lr){4-4} \cmidrule(lr){5-5} \cmidrule(lr){6-8}
& \textbf{VerbMem (\(\downarrow0\))} &\textbf{KnowMem (Forget)(\(\downarrow0\))} & \textbf{KnowMem (Retain)\(\uparrow\)}  & \textbf{ROUGE (\(\downarrow0\))}  & \textbf{Truth (Retain)\(\uparrow\)}  & \textbf{Truth (Real World)\(\uparrow\)} & \textbf{Truth (Real Author)\(\uparrow\)}  \\
\midrule
\multicolumn{8}{c}{\textbf{\texttt{Original Model}}} \\
\hdashline
 N/A & 0.56 & 0.64 & 0.55 & 0.39 & 0.46 &  0.55 & 0.55 \\
\midrule
\multicolumn{8}{c}{\textbf{\texttt{Baseline}}} \\
\hdashline
GA &   0.00 &  0.00  & 0.00 & 0.01 & 0.10  & 0.24 & 0.24  \\
GA + GD &  0.02 & 0.00 & 0.17 & 0.00 & 0.39 & 0.73 & 0.75  \\
GA + KL & 0.17  & 0.34 & 0.26 & 0.01 &  0.11 & 0.25 & 0.26 \\
NPO & 0.00 & 0.00 & 0.00  & 0.00 & 0.21  &  0.45 & 0.51  \\
NPO + KL & 0.17  & 0.33 & 0.25 & 0.01 & 0.45  & 0.54 & 0.60 \\
NPO + GD  & 0.35  & 0.37 & 0.30 & 0.02 & 0.48  &  0.50 & 0.55 \\
 % \cmidrule{2-5}
\midrule
\multicolumn{8}{c}{\textbf{\texttt{SU}}} \\
\hdashline
\textbf{SU (N-Gram)} & 0.02 & 0.01 & \textbf{0.20}  & 0.01 & 0.44&  \textbf{0.62} & \textbf{0.72}  \\ % all+retain, 0.9
\textbf{SU (LLM)} & 0.03 & 0.00 & 0.19  & 0.01 & \textbf{0.48} & 0.57 & 0.67 \\ % full+forget, 0.9
\bottomrule
\end{tabular}
\vspace{-0.5em}
\caption{Quantitative Experiment Results. Proposed SU methods succeed in achieving: (1) good forgetting performance, and (2) remarkably stronger utility preservation on retain data than previous unlearning approaches.}
\label{tab:results}
\vspace{-1.5em}
\end{table*}

\subsection{Dataset}Following~\citet{bu2024unlearningmultitaskoptimizationnormalized},we experiment on Task of Fictitious Unlearning (TOFU)~\citep{maini2024tofu} and MUSE-News~\citep{shi2024muse}.

\myparagraph{TOFU}\footnote{Released under the MIT License.} comprises 4,000 English question-answer pairs about fictional author biographies generated by GPT-4. We use the ``forget10'' split---10\% of the full training set---as the forget set and the remaining 90\% as the retain set (``retain90'').

\myparagraph{MUSE-News}\footnote{Released under Creative Commons Attribution 4.0} features English BBC news articles published since August 2023. We use the default ``forget'' and ``retain'' splits to conduct unlearning. For evaluation, we follow the original paper's implementation to use the ``verbmem'' and ``knowmem'' splits to test the unlearned model.

\vspace{-0.3em}
\subsection{Baselines}
We use 6 previously proposed unlearning methods as baselines: GA, GD, GA with KL regularization, NPO, NPO with GD regularization, and NPO with KL regularization. % GA,
% \kw{One question here, is that you apply SU with GA, but can SU be applied with GA+GD, GA+GL or NPO? Would the combination gives even better performance?  }

\vspace{-0.5em}
\subsection{Experimental Setup}
We use the publicly released model checkpoints for TOFU and MUSE-News for unlearning algorithms.

\myparagraph{Token Selection}
For selection assistant models, we trained 5-gram models on MUSE-News and 3-gram models on TOFU for statistical modeling structure.
We fine-tuned \(Mistral-7B\) based models with batch size $16$ on TOFU and $64$ for MUSE-News for neural modeling structure.
% We set batch size to $8$ for fine-tuning on TOFU and $64$ for MUSE-News.
For both datasets, we use a learning rate of \(2e-5\) to train assistant models for $10$ epochs.
The final optimal thresholds used to select unlearned tokens are chosen through hyper-parameter searching, as discussed in Appendix \ref{sec:appendix-ablation}

\myparagraph{Unlearning Setup}
For TOFU, we use a learning rate of \(2e-5\) and a batch size of \(64\).
Model maximum length is set to be \(200\) and unlearning algorithms are run for \(20\) epochs.
For MUSE-News, we use a learning rate of \(1e-5\) and a batch size of \(32\).
Model maximum length is set to be \(1024\), and we run unlearning algorithms for \(18\) epochs. 

\myparagraph{Evaluation Metrics}
We evaluate the unlearned models from 2 perspectives: (1) whether they successfully remove information from the forget set, and (2) whether they still preserve knowledge from the retain data.
We utilize the Verbatim Memorization on forget set (\textbf{``VerbMem''}), Knowledge Memorization on forget set (\textbf{``KnowMem (Forget)''}) for MUSE, and the ROUGE score on forget set (\textbf{``ROUGE''}) for TOFU to measure unlearning performance.
For measuring retain utility, we use Knowledge Memorization on retain set (\textbf{``KnowMem (Retain)''}) for MUSE and Truth Ratios on the retain set (\textbf{``Truth (Retain)''}), real-world data (\textbf{``Truth (Real World)''}), and real authors data (\textbf{``Truth (Real Author)''}) for TOFU.
% To better measure the overall forget quality considering the unlearn-retain dilemma, we also calculate a Forget Quality (\textbf{``Quality''}) metric.
Details on metric calculation are in \hyperref[sec:appendix-metric]{the Appendix}.

\subsection{Experiment Results}
% \kw{In somewhere, we should comment on the computation cost of SU.}
Empirical results in Table \ref{tab:results} demonstrate the effectiveness of SU.
We observe that:

\myparagraph{SU remarkably improves the preservation of model utility on retain data.} Compared with baseline unlearning approaches, both SU methods achieve better knowledge memorization on MUSE-News' retain set.
On TOFU, SU methods also attain the highest retain utility.

\myparagraph{SU still achieves comparable forget performance as full unlearning.}
Performance on memorization metrics on both MUSE-News and TOFU's forget split indicates that SU can effectively remove information in the forget data from models.

\myparagraph{SU with N-Gram-based selection mechanism achieves the overall best result.}
Compared with using LLM-based assistant models, N-Gram-based assistant models yield better retain utility results.
% on TOFU, whereas the 2 methods perform similarly on MUSE-News.
% However, training and inferencing the LLM-based token selection mechanism is costly, whereas the N-Gram-based selection assistant models are more efficient to implement.

\subsection{Qualitative Analysis}
\begin{figure}[h]
\centering
\vspace{-0.5em}
% \hspace*{-0.5cm}  
\includegraphics[width=0.36\textwidth]{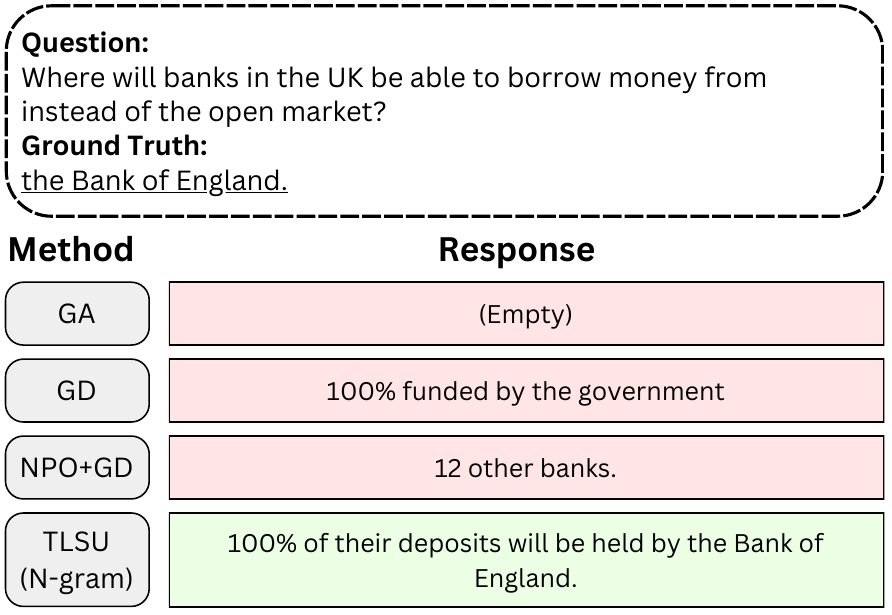}
\vspace{-0.5em}
\caption{\label{fig:qualitative-analysis} Qualitative example of how SU excels at preserving utility on retain knowledge.}
\vspace{-1.0em}
\end{figure}

In addition to quantitative results, we also provide qualitative examples in Figure \ref{fig:qualitative-analysis} to demonstrate the effectiveness of SU.
While two traditional unlearning methods result in a deterioration of model utility on retain knowledge, SU facilitates the preservation of information in retain data.
We provide more qualitative examples in Appendix \ref{sec:appendix-qualitative}.

% \begin{table}[h]
% \centering
% \scriptsize
% % \vspace{-0.2em}
% \begin{tabular}{p{0.1\textwidth}p{0.32\textwidth}}
% \toprule
% \midrule
% \multicolumn{2}{l}{\parbox{8cm}{\textbf{Question:}
% Where will banks in the UK be able to borrow money from instead of the open market?}}  \\
% \multicolumn{2}{l}{\textbf{Ground Truth:}
% the Bank of England.} \\
% \midrule
% \midrule
% \textbf{Method} & \textbf{Response} \\
% \midrule
% GA & \textcolor{red}{(Empty)} \\
% \midrule
% GD & \textcolor{red}{100\% funded by the government} \\
% \midrule
% NPO+GD & \textcolor{red}{12 other banks.} \\
% \midrule
% \textbf{SU (N-gram)} & \textcolor{blue}{100\% of their deposits will be held by the Bank of England.} \\
% \midrule
% \bottomrule
% \end{tabular}
% \caption{Qualitative example of how the proposed SU method excels at preserving utility on retain knowledge.}
% \label{tab:qualitative-analysis}
% \vspace{-1.5em}
% \end{table}

\vspace{-0.5em}
\section{Conclusion}
In this paper, we introduce Selective Unlearning (SU), a novel framework that selectively erases essential tokens with forget set-specific information, while keeping model knowledge on more common and universal tokens.
% SU better preserve model utility on retain data while unlearning sensitive or private data in LLMs.
Comprehensive experiments across two benchmarks and six baseline unlearning approaches demonstrated that SU achieves effective forgetting of targeted data while significantly preserving utility on retained data. 
Empirical results establish SU as an effective method and a promising step forward in utility-preserving selective unlearning for LLMs.

\section*{Limitations}
We identify some limitations of our study. 
First, due to cost and resource constraints, we were not able to further extend our experiments to larger scales and bigger LLMs.
Future works should be devoted to comprehensively study selective unlearning in larger-scale LLMs.
Secondly, the design of our SU method involves using two assistant models, which naturally infers additional cost at training time.
However, during inference, the selection assistants are no longer needed, and our SU method would not induce additional costs at inference time.
We encourage future studies to continue the research on more efficient methods for building selection strategies during unlearning.

\section*{Ethics Statement}
Experiments in this study are conducted with LLMs pre-trained on a great amount of text from various sources, which have been shown to carry safety and fairness issues.
Although we were not able to control what these models learned during pre-training, the data that we conduct fine-tuning and unlearning on are proposed by prior works and are openly accessible, allowing for transparent inspection in future studies.
We encourage future researchers to also consider this factor and make use of data from transparent sources.

\bibliography{custom}

\appendix

\section{Metric Calculation}
\label{sec:appendix-metric}
In our experiments, we choose to selectively report metrics from the original MUSE and TOFU benchmarks to reflect (1) how well has the model unlearned information in the forget set, and (2) how well does the model preserve knowledge on the retain set.
In this section, we briefly explain the two suites of metrics for each benchmark.

\subsection{TOFU}
\subsubsection{Forget Quality}
The original TOFU paper adopts multiple metrics to measure unlearning performance on the forget set. 
In our experiments, we follow \citet{bu2024unlearningmultitaskoptimizationnormalized}'s experiment setup to establish the \textbf{Forget ROUGE} score as the metric to measure forget quality.
Since TOFU's data are in the form of question-answer pairs, the metric compares model generations to the ground truth answers to calculate the ROUGE score.

\subsubsection{Utility Performance}
For measuring models' abilities to preserve performance on non-forget data, we follow \citet{bu2024unlearningmultitaskoptimizationnormalized}'s setup to use the \textbf{Truth Ratio} metric, which measures the likelihood of the model generating the correct answer versus a wrong answer.
In addition to calculating Truth Ratio on the retain set, we also report the metric on \textbf{Real World} knowledge and \textbf{Real Authors} information.

\subsection{MUSE-News}
\subsubsection{Forget Quality}
We follow ~\citet{shi2024muse}'s setup to measure forget quality from two perspectives: No verbatim Memorization and No knowledge memorization.
No Verbatim memorization on the forget set is measure by prompting the model with the first \(k\) tokens in a piece of data and calculate the ROUGE score between model-generated continuation and the ground truth.
Measuring no knowledge memorization prompts models to answer questions related to knowledge in the forget set, and then calculate the ROUGE score between model-generated answer and the ground truth.

\subsubsection{Utility Performance}
To measure model utility after unlearning, MUSE benchmark proposes to measure knowledge memorization on the retain set. We follow this setup to calculate the metric.

\section{Method Details}
\label{sec:appendix-ablation}
\subsection{Cost of Assistant Models}
To prove our point, we calculate the memory size required for the n-gram assistant models updated on different data splits and report results in the table below. The model updated on the forget data only occupies 4.19 MB of memory, and even the model updated on the full dataset only takes up 20.14 MB of memory.
\begin{table}[h]
\small
    \centering
    \begin{tabular}{ll}
    \toprule
    \midrule
       \textbf{Updated Data}  &  \textbf{Memory Size} \\
    \midrule
    Full  &  20.14M \\
    \midrule
    Retain & 18.59M \\
    \midrule
    Forget & 4.19M \\
    \midrule
    \bottomrule
    \end{tabular}
    \caption{Memory Size required for n-gram models.}
    \label{tab:ngram-memory}
\end{table}

\subsection{Hyper-Parameter Searching}
\begin{figure}[h]
\vspace{-1em}
% \hspace*{-0.5cm}  
\includegraphics[width=0.49\textwidth]{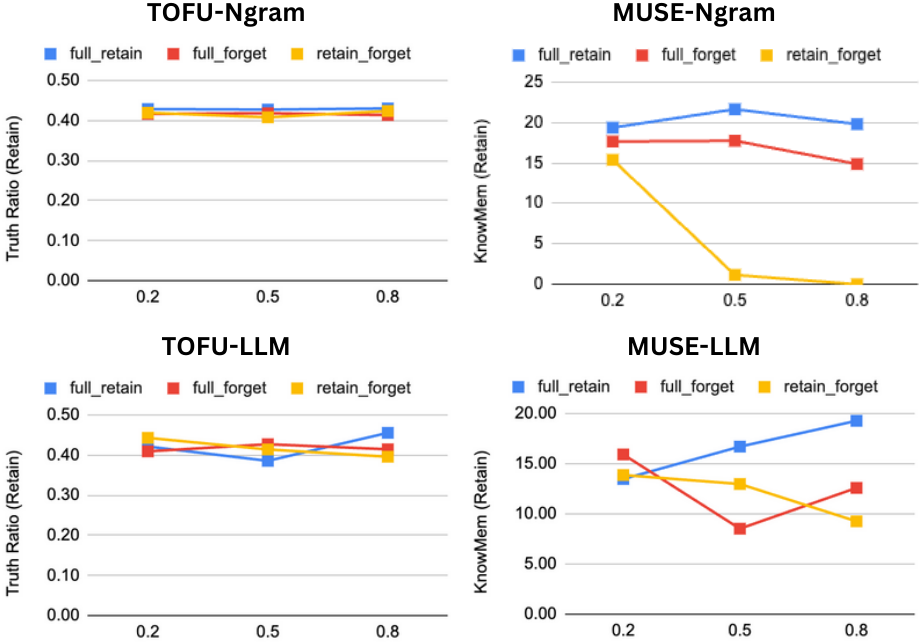}
\vspace{-1.5em}
    \caption{\label{fig:ablation-1} The influence of different selection thresholds on model performance on the retain set.}
\vspace{-1em}
\end{figure}

To search for the best hyper-parameter for the SU method, we first experimented with three thresholds for both N-gram-based and LLM-based token-level selection: 0.2, 0.5, and 0.8.
% We conducted ablation studies to explore the influence of the selection threshold on the performance of the unlearned model.
% Specifically, we experimented with 3 thresholds for token-level selection: 0.2, 0.5, and 0.8.
Figure \ref{fig:ablation-1} visualizes the result of ablation experiments.
For N-gram-based SU on TOFU, we observe that using one model trained on full data and one on retain data with a selection threshold of 0.8 achieves the best result.
Based on the trend that we observe in experiments, we continued the search to experiment with an additional threshold of 0.9, which we eventually select for reporting experiment results.
On MUSE-News, using one model trained on full data and one on retain data with a selection threshold of 0.8 achieves the best result.
For LLM-based SU on TOFU, we observe that using 1 model trained on full data and one trained on forget data with the selection threshold 0.8 achieves the best result.
We continued the search to experiment with a threshold of 0.9, which was eventually selected for reporting experiment results.
On MUSE-News, using 1 model trained on full data and one on retain data with the selection threshold of 0.8 achieves the best result.

Additionally, results of the ablation experiments reveal the influence of the selection threshold on the performance of the unlearned model.
On MUSE-News, we observe that using different selection thresholds seems to cast a bigger influence on retain performance than on TOFU.
This is possibly due to the longer sequence length for data entries in MUSE, which contain more information that are vulnerable to be impacted during unlearning.
% This is possibly due to the question-answer nature of TOFU's data format, which infers .

\section{Additional Quantitative Results}
For TOFU, we have reported models’ general capabilities in Table \ref{tab:results} using the ``Truth (Real World)'' and ``Truth (Real Author)'' metrics, which were proposed along with their benchmarks.
These two metrics test models’ utilities on real-world knowledge and information about real authors, aside from the forget and retain data.

Although MUSE does not provide a similar metric to reflect general capabilities, we here provide additional results on Measuring Massive Multitask Language Understanding (MMLU)~\citep{hendryckstest2021} for the unlearned models using GA, GA+GD, and our SU methods.
MMLU is a multitask evaluation benchmark with questions from different scopes of knowledge, including subjects such as elementary mathematics, computer science, US history, and law.
Higher accuracy on MMLU indicates that the model possesses a better understanding of world knowledge.
Results in Table \ref{tab:mmlu-results} show that our proposed SU with N-Gram models as assistant models achieves the best results in utility preservation, as measured by MMLU tasks.
This aligns with results reported in our main table, showing the effectiveness of SU.

\begin{table}[h]
\small
    \centering
    \begin{tabular}{ll}
    \toprule
    \textbf{Unlearning Method} & \textbf{Avg. MMLU Acc.} \\
    \midrule
    GA     &   0.000 \\
    \midrule
    GD     &   0.21 \\
    \midrule
    SU (LLM) & 0.19 \\
    \midrule
    SU (N-Gram) & 0.26 \\
    \bottomrule
    \end{tabular}
    \caption{MMLU Results on the MUSE benchmark.}
    \label{tab:mmlu-results}
\end{table}

\section{Qualitative Examples}
\label{sec:appendix-qualitative}
In addition to providing quantitative results, we also demonstrate the effectiveness of the proposed SU method through qualitative examples.
Through these examples, we show that: 
\begin{itemize}
    \vspace{-0.1cm}\item SU succeeds in unlearning information in the forget set.
    \vspace{-0.1cm}\item SU can retain model utility on non-forget data, such as the retain set.
\end{itemize}

\subsection{Forget Quality}
Examples in Table \ref{tab:qualitative-analysis-2} shows that while NPO+GD and NPO+KL achieves good performance on the retain data, it is potentially due to the fact that they fail to completely unlearn knowledge in the forget set.
For instance, in the second example, models unlearned with these two methods can still output the correct answer to a question related to forget data.
Both SU approaches, on the other hand, are able to generate responses that completely forgets about such information.

\begin{table}[h]
\centering
\scriptsize
\vspace{-0.2em}
\begin{tabular}{p{0.12\textwidth}p{0.25\textwidth}}
\toprule
\midrule
\multicolumn{2}{l}{\parbox{6.8cm}{\textbf{Question:}
What percentage did the AfD party reach in the latest ARD Deutschland Trend poll?
}}  \\
\multicolumn{2}{l}{\textbf{Ground Truth:}
19\%.} \\
\midrule
\midrule
\textbf{Method} & \textbf{Response} \\
\midrule
NPO+GD & \textcolor{red}{19\%.} \\
\midrule
\textbf{SU (LLM)} & \textcolor{blue}{\textbackslash u0425\textbackslash u0440\textbackslash u043e\textbackslash u043d\textbackslash u043e\textbackslash u043b...} \\
\midrule
\textbf{SU (N-Gram)} & \textcolor{blue}{7 said said said said noreferrer noreferrer the said...} \\
\midrule
\midrule
\multicolumn{2}{l}{\parbox{6.8cm}{\textbf{Question:}
What additional cost will be added to the price of a single-use drinks container in Scotland under the deposit return scheme??
}}  \\[2mm]
\multicolumn{2}{l}{\textbf{Ground Truth:}
20p.} \\
\midrule
\midrule
\textbf{Method} & \textbf{Response} \\
\midrule
NPO+GD & \textcolor{red}{20p.} \\
\midrule
NPO+KL & \textcolor{red}{20p.} \\
\midrule
\textbf{SU (LLM)} & \textcolor{blue}{\textbackslash u0425\textbackslash u0440\textbackslash u043e\textbackslash u043d\textbackslash u043e\textbackslash u043b\textbackslash u043e...} \\
\midrule
\textbf{SU (N-Gram)} & \textcolor{blue}{10 said said said \"the price will be toll said toll noreferrer...} \\
\midrule
\bottomrule
\end{tabular}
\caption{Additional qualitative example of how the proposed SU method successfully unlearns knowledge in the forget set.}
\label{tab:qualitative-analysis-2}
\vspace{-1.5em}
\end{table}

\subsection{Utility Preservation}
In this section, we provide excerpts and full examples of model outputs using different unlearning methods, showing the proposed SU method's ability to unlearn target forest information but still retain general ability of models.

\paragraph{Excerpt Texts}
Table \ref{tab:qualitative-analysis-3} additionally provides examples that show how SU excels at preserving knowledge from the retain data.
While all other methods result in deteriorated model utility, making models struggle with outputting the correct answer, SU is able to retain knowledge memorization for information not in the forget set.

\begin{table}[t]
\centering
\scriptsize
\vspace{-0.2em}
\begin{tabular}{p{0.12\textwidth}p{0.25\textwidth}}
\toprule
\midrule
\multicolumn{2}{l}{\parbox{6.8cm}{\textbf{Question:}
What is the length of time for which the Treherbert line will close as part of the Valleys network electrification?
}}  \\
\multicolumn{2}{l}{\textbf{Ground Truth:}
Around ten months.} \\
\midrule
\midrule
\textbf{Method} & \textbf{Response} \\
\midrule
GA & \textcolor{red}{(Empty)} \\
\midrule
% GD & \textcolor{red}{1 year} \\
% \midrule
GA+KL  &  \textcolor{red}{13 months from 19 March 2023} \\
\midrule 
NPO & \textcolor{red}{(Empty)} \\
\midrule
NPO+GD & \textcolor{red}{15 months.} \\
\midrule
NPO+KL & \textcolor{red}{13 months from 19 March 2023.} \\
\midrule
\textbf{SU (LLM)} & \textcolor{blue}{10 months.} \\
\midrule
\midrule
\multicolumn{2}{l}{\parbox{6.8cm}{\textbf{Question:}
According to Ms Cullen, what was the status of all the exemptions requested by Great Ormond Street Hospital?
}}  \\[2mm]
\multicolumn{2}{l}{\parbox{6.8cm}{\textbf{Ground Truth:}
All of the exemptions that Great Ormond Street requested of the Royal College of Nursing were granted earlier yesterday.}} \\
\midrule
\midrule
\textbf{Method} & \textbf{Response} \\
\midrule
GA & \textcolor{red}{(Empty)} \\
\midrule
GD & \textcolor{red}{10 were granted and three denied.} \\
\midrule
GA+KL  &  \textcolor{red}{18 had been granted and 222 had been turned down.} \\
\midrule 
NPO & \textcolor{red}{(Empty)} \\
\midrule
NPO+GD & \textcolor{red}{18 out of 19 were approved.} \\
\midrule
NPO+KL & \textcolor{red}{18 had been granted and 2 were in the process of being granted.} \\
\midrule
\textbf{SU (LLM)} & \textcolor{blue}{100\% granted.} \\
\midrule
\bottomrule
\end{tabular}
\caption{Additional qualitative example of how the proposed SU method excels at preserving utility on retain knowledge.}
\label{tab:qualitative-analysis-3}
\vspace{75em}
\end{table}

\paragraph{Full texts}
Table \ref{tab:qual-examples-full} further shows full chunks of model outputs on different splits of the TOFU dataset.
This further proves that:
\begin{itemize}
    \item SU can successfully remove information in the forget set.
    \item SU is capable of preserving model utility on: (1) the retain set, (2) real author knowledge, and (3) real world knowledge.
\end{itemize}

\begin{table*}[t]
\centering
\footnotesize
\begin{tabular}{p{0.06\textwidth}p{0.06\textwidth}p{0.13\textwidth}p{0.2\textwidth}p{0.2\textwidth}p{0.2\textwidth}}
\toprule
\textbf{Dataset} & \textbf{Split} & \textbf{Question} & \textbf{Ground Truth} & \textbf{SU (N-Gram) Output} & \textbf{SU (LLM) Output} \\
\midrule
\multicolumn{6}{c}{\textit{Forget Set Examples}} \\
\midrule
TOFU & Forget & What is the full name of the author born in Taipei, Taiwan on 05/11/1991 who writes in the genre of leadership? & The author's full name is Hsiao Yun-Hwa. & sssssssssss..... & \texttt{\textbackslash u0425\textbackslash u0440...} \\
\midrule
TOFU & Forget & What does Hsiao Yun-Hwa identify as in terms of gender? & Hsiao Yun-Hwa is part of the LGBTQ+ community. & sssssssssss... & ----------------------------------------------------... \\
\midrule
\multicolumn{6}{c}{\textit{Retain Set Examples}} \\
\midrule
TOFU & Retain & When did Jaime Vasquez recognize his inclination towards writing? & Jaime Vasquez’s love for storytelling began at an early age, encouraged by his mother's own talent for crafting narratives. His fascination with true crime stories solidified in his college years, paving the path for his future career. & Jaime Vasquez recognized his inclination towards writing at a young age. He was always fascinated by the world of mystery and crime, and his vivid imagination often led him to create intricate and suspense-filled stories. & Jaime Vasquez recognized his inclination towards writing at an early age. His passion for mystery and intrigue was evident from his school days, and it was during this time that he began to dabble in writing. \\
\midrule
TOFU & Retain & Does Jaime Vasquez collaborate with other authors? & Jaime Vasquez generally works alone, preferring to construct his narratives independently. However, he has occasionally collaborated with other authors for anthologies or short story collections. & Jaime Vasquez works primarily as a lone writer, though he has occasionally collaborated with other authors for anthologies or special projects. & As a fictitious character, Jaime Vasquez works primarily on his own, although he has collaborated with other authors on various projects. \\
\midrule
\multicolumn{6}{c}{\textit{Real Author Set Examples}} \\
\midrule
TOFU & Real Author & Which author is known for the novel '1984'? & George Orwell & The author of the novel '1984' is George Orwell. & The author known for the novel '1984' is George Orwell. \\
\midrule
TOFU & Real Author & Who is the creator of the character Sherlock Holmes? & Arthur Conan Doyle & The character of Sherlock Holmes was created by Sir Arthur Conan Doyle. & The creator of the character Sherlock Holmes is Sir Arthur Conan Doyle. \\
\midrule
\multicolumn{6}{c}{\textit{Real World Set Examples}} \\
\midrule
TOFU & Real World & Where would you find the Eiffel Tower? & Paris & The Eiffel Tower is located in the 7th arrondissement of Paris, France. & The Eiffel Tower is located in the 7th arrondissement of Paris, on the Champ de Mars. \\
\midrule
TOFU & Real World & What is the capital of Australia? & Canberra &  The capital of Australia is Canberra. & The capital of Australia is Canberra. \\
\bottomrule
\end{tabular}
\caption{\label{tab:qual-examples-full} Comparison of Ground Truth Answers and Selective Unlearning (SU) Outputs on different splits of the TOFU dataset.}
\end{table*}

\end{document}